\documentclass[runningheads]{llncs}
\usepackage[T1]{fontenc}
\usepackage{amsmath}
\usepackage{bm}
\usepackage{dsfont}

\usepackage{graphicx}
\usepackage{cleveref}

\usepackage{color}

\usepackage{caption}
\usepackage{subcaption}

\usepackage{tabularx,booktabs}



\usepackage[textsize=footnotesize,textwidth=4cm]{todonotes}
\let\oldtodo\todo
\renewcommand{\todo}[1]{\oldtodo{#1}\PackageWarning{TODO:}{TODO: #1}}

\begin{document}
\title{Self-Supervised \\Radiograph Anatomical Region Classification -- How Clean Is Your Real-World Data?}
\titlerunning{Self-Supervised Radiograph Anatomical Region Classification}
%

\author{Simon Langer\inst{1} \and Jessica Ritter\inst{2} \and Rickmer Braren\inst{2} \and Daniel~Rueckert\inst{1,3} \and Paul~Hager\inst{1,2}}

\authorrunning{S. Langer et al.}

%

\institute{Lab for AI in Medicine, Technical University of Munich, Munich, Germany \and
Klinikum rechts der Isar, Technical University of Munich, Munich, Germany \and
Department of Computing, Imperial College London, London, United Kingdom \\
\email{{simon.langer@tum.de}} }

\maketitle              
\begin{abstract}
Modern deep learning-based clinical imaging workflows rely on accurate labels of the examined anatomical region. Knowing the anatomical region is required to select applicable downstream models and to effectively generate cohorts of high quality data for future medical and machine learning research efforts. However, this information may not be available in externally sourced data or generally contain data entry errors. To address this problem, we show the effectiveness of self-supervised methods such as SimCLR and BYOL as well as supervised contrastive deep learning methods in assigning one of 14 anatomical region classes in our in-house dataset of 48,434 skeletal radiographs. We achieve a strong linear evaluation accuracy of 96.6\% with a single model and 97.7\% using an ensemble approach. Furthermore, only a few labeled instances (1\% of the training set) suffice to achieve an accuracy of 92.2\%, enabling usage in low-label and thus low-resource scenarios. Our model can be used to correct data entry mistakes: a follow-up analysis of the test set errors of our best-performing single model by an expert radiologist identified 35\% incorrect labels and 11\% out-of-domain images. When accounted for, the radiograph anatomical region labelling performance increased  -- without and with an ensemble, respectively -- to a theoretical accuracy of 98.0\% and 98.8\%.

\end{abstract}
\section{Introduction}
Radiography is a prominent technique in modern medical diagnostics, capturing approximately 36.4 million images in 2018 in Germany alone \cite{XRayBFS}. The widespread adoption of PACS (Picture Archiving and Communication Systems) makes the utilization of this vast pool of data for research and diagnostics easier; however, even some of the most basic metadata, such as the captured anatomical region, may be missing in the DICOM (Digital Imaging and Communications in Medicine) files when recorded at external hospitals, or be generally unreliable \cite{Gueld2002QualityOD}. To effectively utilize this data \cite{Larson2020EthicsOU}, we show how to use self-supervised methods for the accurate prediction of anatomical regions of radiograph images. This is particularly useful when few annotations are available, allowing our approach to be used by hospitals with limited technological and personnel resources \cite{Tinto2013GoodCP,Willett2018ImagingIT}.

\begin{figure}[t]
    \centering
    \includegraphics[width=\textwidth]{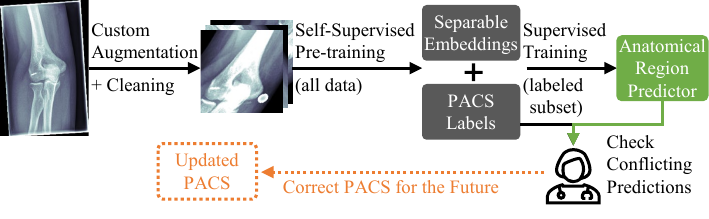}
    \caption{Overview -- we pre-train our backbone using self-supervision, then train a fully connected head, and finally use its high quality predictions to correct noisy PACS labels.}
    \label{fig:Approach}
\end{figure}

Identifying the correct anatomical region of new or existing data has several important applications: 
\textsc{i)} It increases the data quality in an existing PACS system, detecting outliers or incorrect labels by comparing predictions with the existing anatomical region metadata. We explore this application in \cref{subsec:Data Quality Assessment}.
\textsc{ii)} It increases dataset sizes for research by integrating external data (frequently missing the crucial anatomical metadata) automatically.
\textsc{iii)} It improves downstream application reliability by employing the predictions as a means of verifying the input for models which are only intended to be applied to certain anatomical regions. This avoids erroneous results and other out-of-domain misbehavior.
\textsc{iv)} It aids underfunded hospitals during the establishment of their PACS systems by reducing the amount of metadata that must be manually annotated. \Cref{subsec:Self-Supervision excels in a low data regime} demonstrates the strength of our self-supervised approach in such a low data regime.
Our contributions (overview in \cref{fig:Approach}) are:

\begin{itemize}
    \item a high-performance self-supervised anatomical region classification architecture whose strength we show on an in-house dataset of over 48,000 skeletal radiographs with 14 classes, achieving an ensemble accuracy of 97.7\%
    \item custom data augmentations and cleaning which enhance plausibility by reducing the model's focus on operation planning gauges and image borders, verified via Grad-CAM explanations
    \item an analysis of our model's results by an expert radiologist, highlighting its effectiveness in detecting incorrect PACS labels
\end{itemize}

\section{Related Work}
As the cost of creating high quality annotations for medical data is very high, methods capable of extracting knowledge without labels have been of major interest in the medical deep learning community.

Self-supervised clustering methods such as DeepCluster \cite{Caron2018DeepCF} exploit the powerful prior of local information in CNNs (Convolutional Neural Networks). Alternating between generating features from the CNN and clustering said features using k-means \cite{MacQueen1967SomeMF}, a pre-chosen number of clusters is generated. 

\label{sec:Related Work:SimCLRBYOL}Other self-supervised methods learn to generate image embeddings through a context matching task which computes features that uniquely describe a given image, being able to re-identify it despite heavy augmentations such as crops, rotations and intensity shifts. SimCLR (a Simple framework for Contrastive Learning of visual Representations) \cite{Chen2020ASF} is a contrastive method which employs both positive and negative examples for this purpose, while BYOL (Bootstrap Your Own Latent) \cite{Grill2020BootstrapYO} predicts the feature representation of its target network from an augmented image encoded by the online network. More details on SimCLR and BYOL are available in \cref{subsec:Architecture and (Pre)training Objectives}. The resulting encoder can be used as a backbone for downstream image analysis tasks.

The task of supervised anatomical region classification has been previously attempted with varying selections of regions and radiography images \cite{Dratsch2020PracticalAO,Jonske2022DeepLC}; e.g.,~Fang et al.~use supervised deep learning to predict anatomical regions at multiple granularities \cite{Fang2020GeneralizedRV}. In contrast, DeepMCAT \cite{Kart2021DeepMCAT} applies DeepCluster \cite{Caron2018DeepCF} to classify different Cardiac MR views (Magnetic Resonance Imaging). Most closely related to our work, SAM-X \cite{Hinterwimmer2022SAMXSA} employs DeepCluster \cite{Caron2018DeepCF} to generate 1000 clusters of images from a musculoskeletal tumor center. The clusters are then manually annotated, yielding weak semantic labels or the decision to discard the cluster. These weak labels were merged to generate 28 skeletal classes that were used as subsequent targets for the CNN optimized by DeepCluster \cite{Caron2018DeepCF}. The main two drawbacks of this approach are \textsc{a)} heavy reliance on the assumption that each of the clusters is homogeneous w.r.t.~the target class (i.e.~anatomical region), and \textsc{b)} the need for explicit human interaction to determine the predominant (and hopefully only) anatomical region. We address both issues with this work as our architecture directly assigns each radiograph to an anatomical region using only those labels generated during normal clinical practice.

\section{Dataset and Methodology}

\subsection{Dataset Generation}
14 anatomical regions that frequently appear in clinical practice were defined as our classification targets. A total of 48,434 radiographs with corresponding anatomical region labels from our hospital PACS were exported as DICOM files. The associated anatomical region information was used as our initial, yet noisy (see \cref{subsec:Data Quality Assessment}) ground truth labels. Subsequently, we split the dataset into 31,011 (64\%)/7,677 (16\%)/9,746 (20\%) training/validation/test images (see supplementary material for per-class distributions).

\subsection{Augmentation and Image Cleaning}\label{subsec:Augmentation and Cleaning}
As we work with unfiltered real-world medical data, we encounter cases such as arbitrarily rotated radiographs which frequently contain a white border around the perimeter of the original capture image. We detect and remove the border, and normalize the rotation using OpenCV \cite{opencv_library}.
Furthermore, images in the region pelvis/hip and rarely knee can contain a circular gauge for surgery planning. To combat the model relying on this for its prediction (see \cref{subsec:Custom Augmentations and Plausibility}), we add a novel augmentation where we use 6 example gauges extracted from the dataset, and randomly insert zero to two of them at random locations and scales in all images. 
Moreover, we apply the standard SimCLR augmentations color jitter, random affine and random crop resize (for hyperparameter details and visualizations, see the supplementary material). To preserve fine-grained features of the medical images, the Gaussian blur augmentation was not used \cite{Azizi2021BigSM}.

\subsection{Architecture and Pre-training Objectives}\label{subsec:Architecture and (Pre)training Objectives}
Serving as the backbone of our model architecture (i.e. self-supervised feature extractor), we use PyTorch's \cite{Paszke2019PyTorchAI} standard implementation of a randomly initialized ResNet18 \cite{He2015DeepRL} and remove the final fully connected layer, resulting in embeddings of size 512 after global average pooling. An initial comparison with ResNet50 showed no benefit from using a larger model for our task. We perform our pretrainings using the Adam optimizer \cite{DBLP:journals/corr/KingmaB14}, a weight decay of $10^{-4}$, a learning rate of $3\cdot 10^{-4}$ (with cosine annealing \cite{Loshchilov2016SGDRSG}), and train for 1000 epochs.

\subsubsection{SimCLR}\label{subsubsec:SimCLR}
(\cite{Chen2020ASF}), a contrastive approach, processes batches of positive and negative pairs: positive pairs
 are two differently augmented views of the same image ($i,j$), while negative pairs are from two different source images. The subsequent projection head consists of a 2-layer MLP (multilayer perceptron) with ReLU (Rectified Linear Unit) activations and BatchNorm \cite{Ioffe2015BatchNA}, computing the projections $\bm{z}_i,\bm{z}_j$. These are passed to the normalized temperature ($\tau$) scaled cross entropy, which compares the cosine similarity ($\mathrm{sim}$) of positive/negative pairs:

\begin{equation}\label{eq:SimCLR}
    \mathcal{L}_{i,j}=-\log\frac{\exp(\mathrm{sim}(\bm{z}_i,\bm{z}_j)/\tau)}{\sum_{k=1}^{2N} \mathds{1}_{[k\neq{}i]}\exp(\mathrm{sim}(\bm{z}_i,\bm{z}_k)/\tau)}
\end{equation}

We set $\tau=0.5$ (as suggested in \cite{Chen2020ASF}) and train using a batch size of 1024.

\subsubsection{BYOL}\label{subsubsec:BYOL}
(\cite{Grill2020BootstrapYO}) does not rely on negative pairs; instead, it uses an online and target network with separate weights $\theta$ and $\xi$. After the projection to $z_\theta$ and $z_\xi'$, the online network attempts to predict the target projection by computing $q_\theta(z_\theta)$.
No gradient information is backpropagated through the target network. Prior to computing the mean squared error loss (MSE), both prediction and target are $L_2$ normalized. Finally, the target network is updated as an exponential moving average of the online network's parameters.

Due to VRAM constraints, we had to reduce our batch size to 896 for BYOL; accordingly, we set $\tau_{base}=0.9995$ as recommended in \cite{Grill2020BootstrapYO}.

\subsubsection{Supervised Contrastive Learning}
As label information is available for the whole dataset, we also use supervised contrastive learning \cite{Khosla2020SupervisedCL} for one of our models, extending the positive pairs of SimCLR to all views which have the same downstream class. 

\subsubsection{Evaluation}\label{subsubsec:Evaluation}
Following the standard linear evaluation protocol, a single fully connected layer is placed after the frozen pretrained backbone to directly predict the 14 anatomical region classes. We train using the cross-entropy loss, Adam, no weight decay, and a learning rate of $5\cdot 10^{-2}$ (with cosine annealing). We recommend not to unfreeze the backbone to allow for future evaluation of the training and validation data for misclassified regions (see \cref{subsec:Data Quality Assessment}).

\section{Results and Discussion}\label{sec:Results and Discussion}
Our first evaluation results summarize both the per anatomical region and overall performances as top-1 accuracy on our test set when training on our entire training set (\cref{tab:maintable}). We achieve excellent accuracy both on average and for the individual classes, with all approaches achieving over $96.5\%$ mean accuracy. Overall, this shows the strength of the features learned through the contrastive pre-training process. Note the slightly lower scores for the spine classes -- our investigations in \cref{subsec:Data Quality Assessment} tackle the underlying issue: noisy PACS labels.%

\begin{table}[tb]
\centering
\caption{Top-1 test accuracy of our three training approaches (employing the entire training set); spine abbreviated as `s'.}
\label{tab:maintable}
\begin{tabular}{@{}lrrrrrrrr@{}}
\toprule
\textbf{Approach} &
  \textbf{ALL} &
  ~~~\begin{tabular}[c]{@{}r@{}}clavicle\\      shoulder\end{tabular} &
  ~~~\begin{tabular}[c]{@{}r@{}}skull\\      rib\end{tabular} &
  ~~~\begin{tabular}[c]{@{}r@{}}elbow\\      knee\end{tabular} &
  ~~~\begin{tabular}[c]{@{}r@{}}wrist\\      hand\end{tabular} &
  ~~~\begin{tabular}[c]{@{}r@{}}foot\\      ankle\end{tabular} &
  ~\begin{tabular}[c]{@{}r@{}}pelvis/hip\\      cervical s\end{tabular} &
  ~\begin{tabular}[c]{@{}r@{}}thoracic s\\      lumbar s\end{tabular} \\ \midrule
\textbf{SimCLR} &
  96.6\% &
  \begin{tabular}[c]{@{}r@{}}98.0\%\\      97.9\%\end{tabular} &
  \begin{tabular}[c]{@{}r@{}}96.8\%\\      97.1\%\end{tabular} &
  \begin{tabular}[c]{@{}r@{}}96.4\%\\      97.5\%\end{tabular} &
  \begin{tabular}[c]{@{}r@{}}97.0\%\\      98.5\%\end{tabular} &
  \begin{tabular}[c]{@{}r@{}}97.4\%\\      98.1\%\end{tabular} &
  \begin{tabular}[c]{@{}r@{}}97.2\%\\      95.5\%\end{tabular} &
  \begin{tabular}[c]{@{}r@{}}90.3\%\\      94.1\%\end{tabular} \\ \cmidrule(l){3-9} 
\textbf{BYOL} &
  96.6\% &
  \begin{tabular}[c]{@{}r@{}}98.6\%\\      97.1\%\end{tabular} &
  \begin{tabular}[c]{@{}r@{}}94.3\%\\      97.1\%\end{tabular} &
  \begin{tabular}[c]{@{}r@{}}96.0\%\\      98.6\%\end{tabular} &
  \begin{tabular}[c]{@{}r@{}}97.5\%\\      98.7\%\end{tabular} &
  \begin{tabular}[c]{@{}r@{}}97.9\%\\      98.7\%\end{tabular} &
  \begin{tabular}[c]{@{}r@{}}97.2\%\\      93.2\%\end{tabular} &
  \begin{tabular}[c]{@{}r@{}}90.6\%\\      94.8\%\end{tabular} \\ \cmidrule(l){3-9} 
\textbf{SupCon} &
  97.1\% &
  \begin{tabular}[c]{@{}r@{}}98.4\%\\      98.5\%\end{tabular} &
  \begin{tabular}[c]{@{}r@{}}97.5\%\\      97.6\%\end{tabular} &
  \begin{tabular}[c]{@{}r@{}}97.8\%\\      99.0\%\end{tabular} &
  \begin{tabular}[c]{@{}r@{}}97.5\%\\      98.6\%\end{tabular} &
  \begin{tabular}[c]{@{}r@{}}96.6\%\\      98.1\%\end{tabular} &
  \begin{tabular}[c]{@{}r@{}}94.0\%\\      95.0\%\end{tabular} &
  \begin{tabular}[c]{@{}r@{}}92.4\%\\      93.8\%\end{tabular} \\ \bottomrule
\end{tabular}
\end{table}

\begin{figure}[tb]
    \centering
    
    {
    \includegraphics[width=0.58\textwidth]{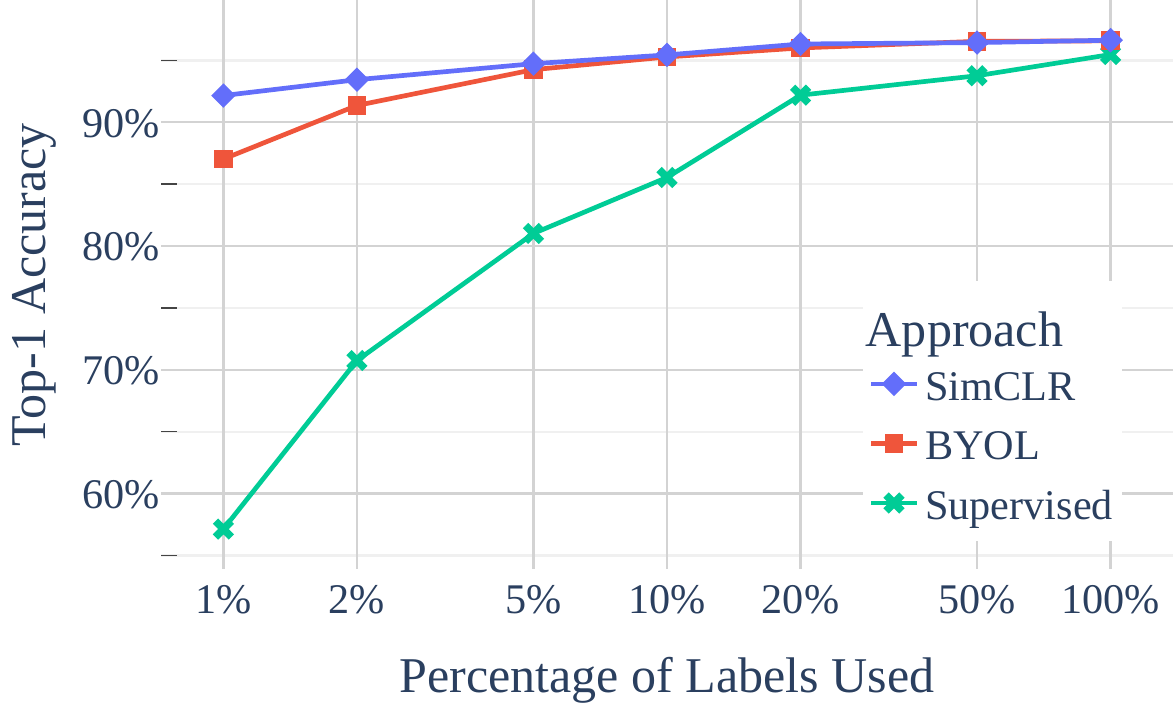}
    }
    \caption{Impact of the amount of labeled data on final performance. This shows the importance of using self-supervised methods in low-label (i.e. low-resource) settings, as the supervised baseline is outperformed by a large margin by our self-supervised approaches, which already perform very well at $\geq 1\%$ of all training data. Note that the x-axis is scaled logarithmically. }
    \label{fig:LowData}
\end{figure}

\begin{figure}[t!]
    \centering
    \newcommand{\curfigwidth}{0.26}
    \begin{subfigure}[b]{\curfigwidth\textwidth}
        \includegraphics[width=\textwidth]{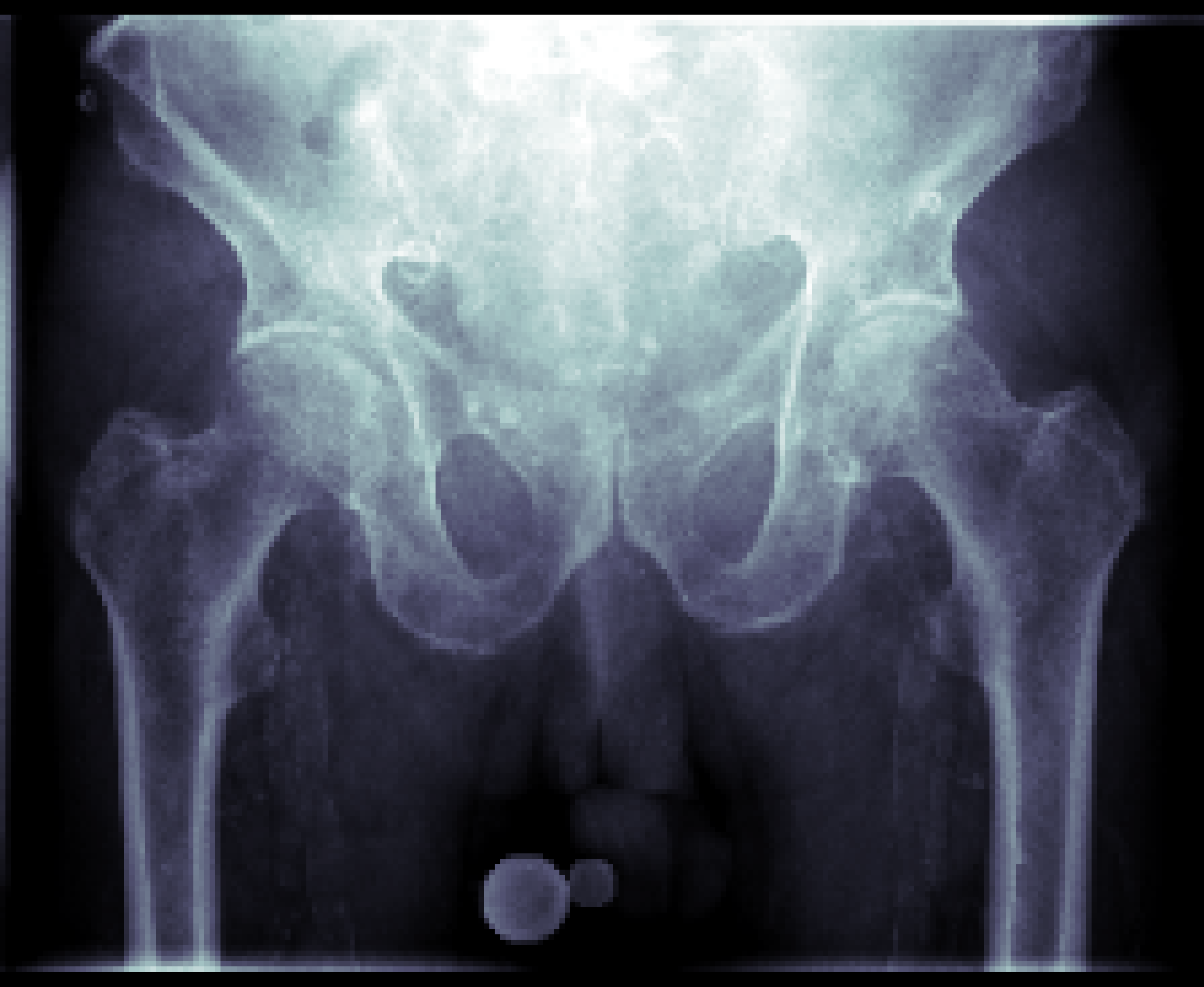}
        \includegraphics[width=\textwidth]{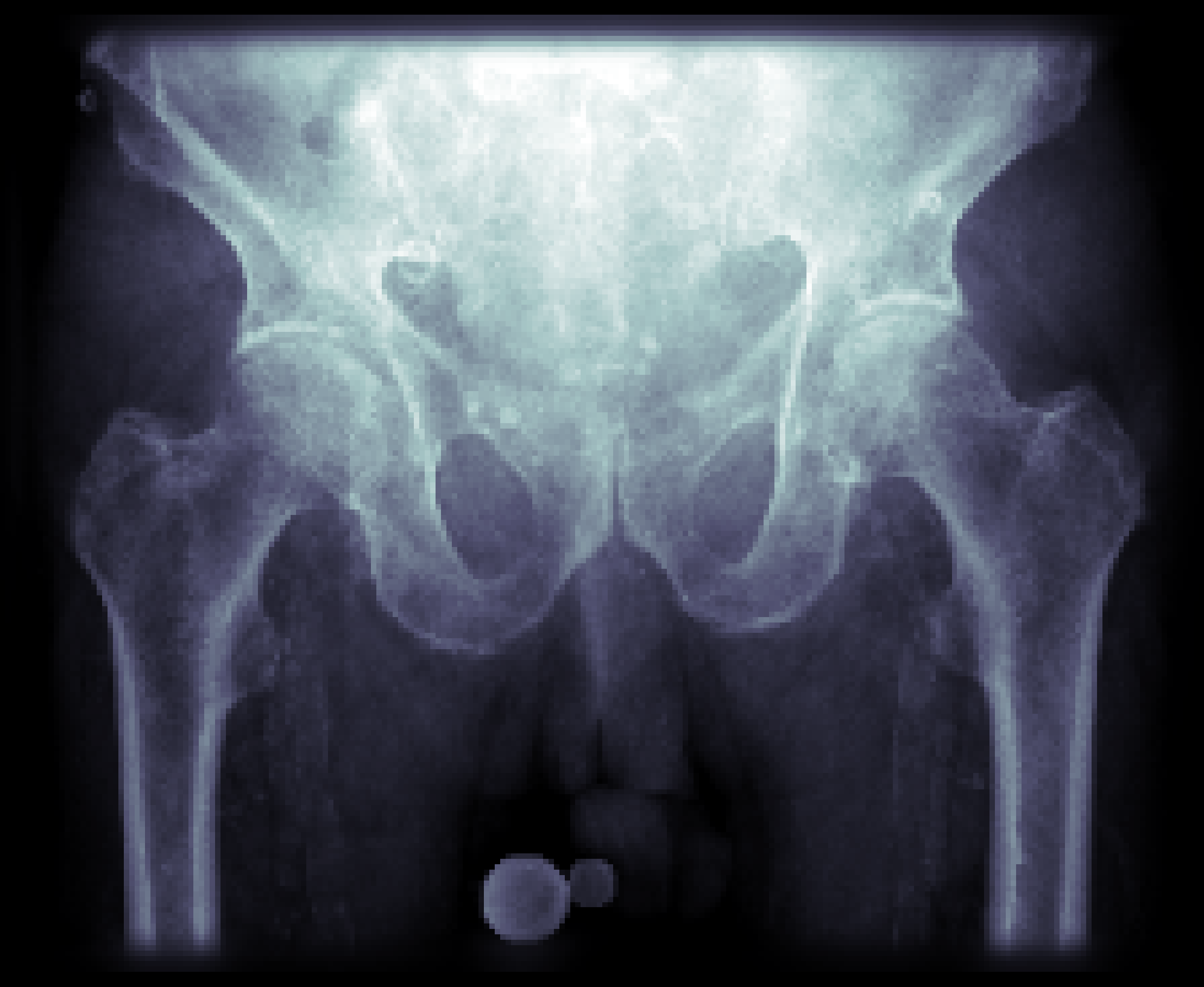}
        \caption{Model Input}
        \label{subfig:GradCam:Raw}
    \end{subfigure}
    \begin{subfigure}[b]{\curfigwidth\textwidth}
        \includegraphics[width=\textwidth]{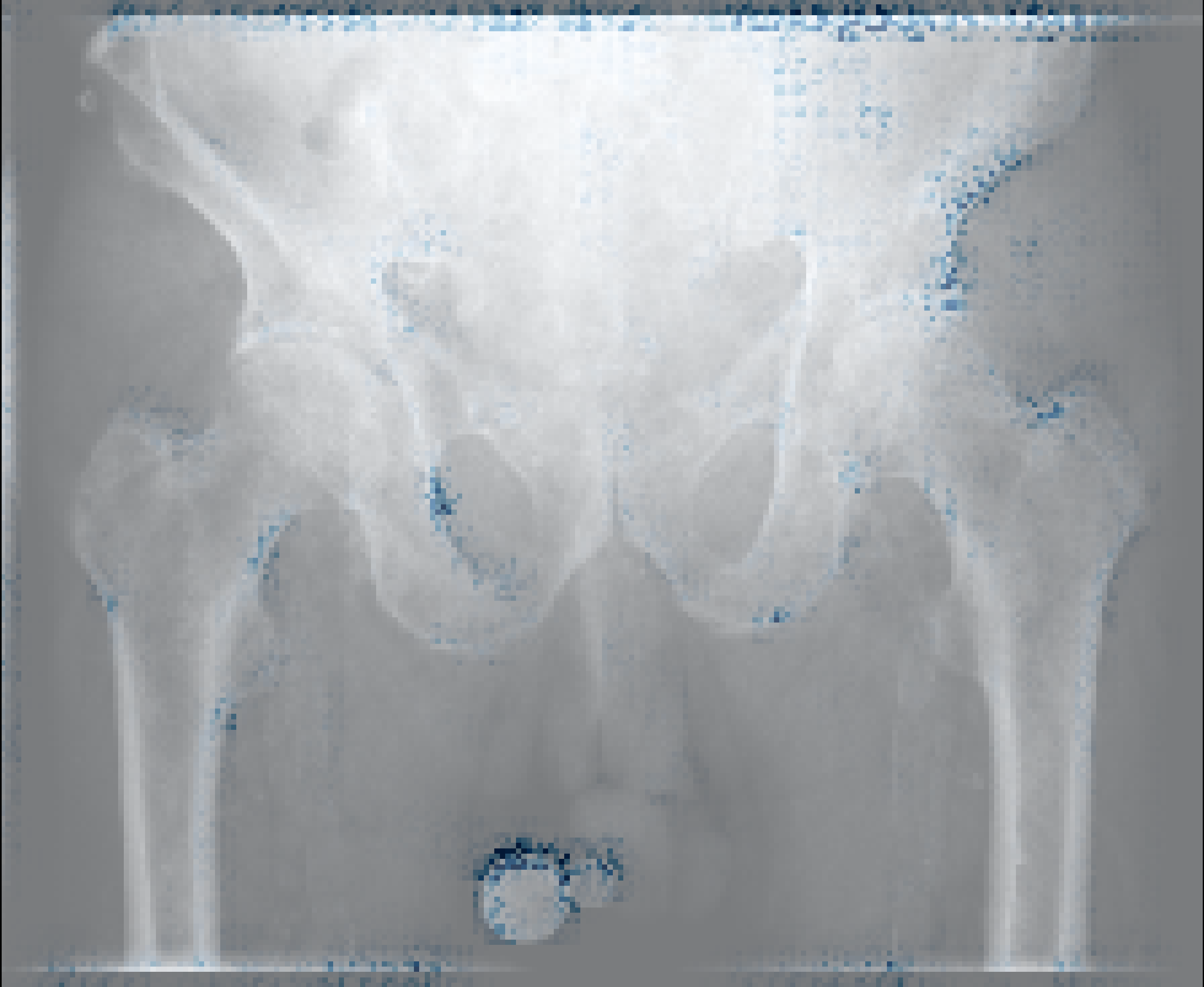}
        \includegraphics[width=\textwidth]{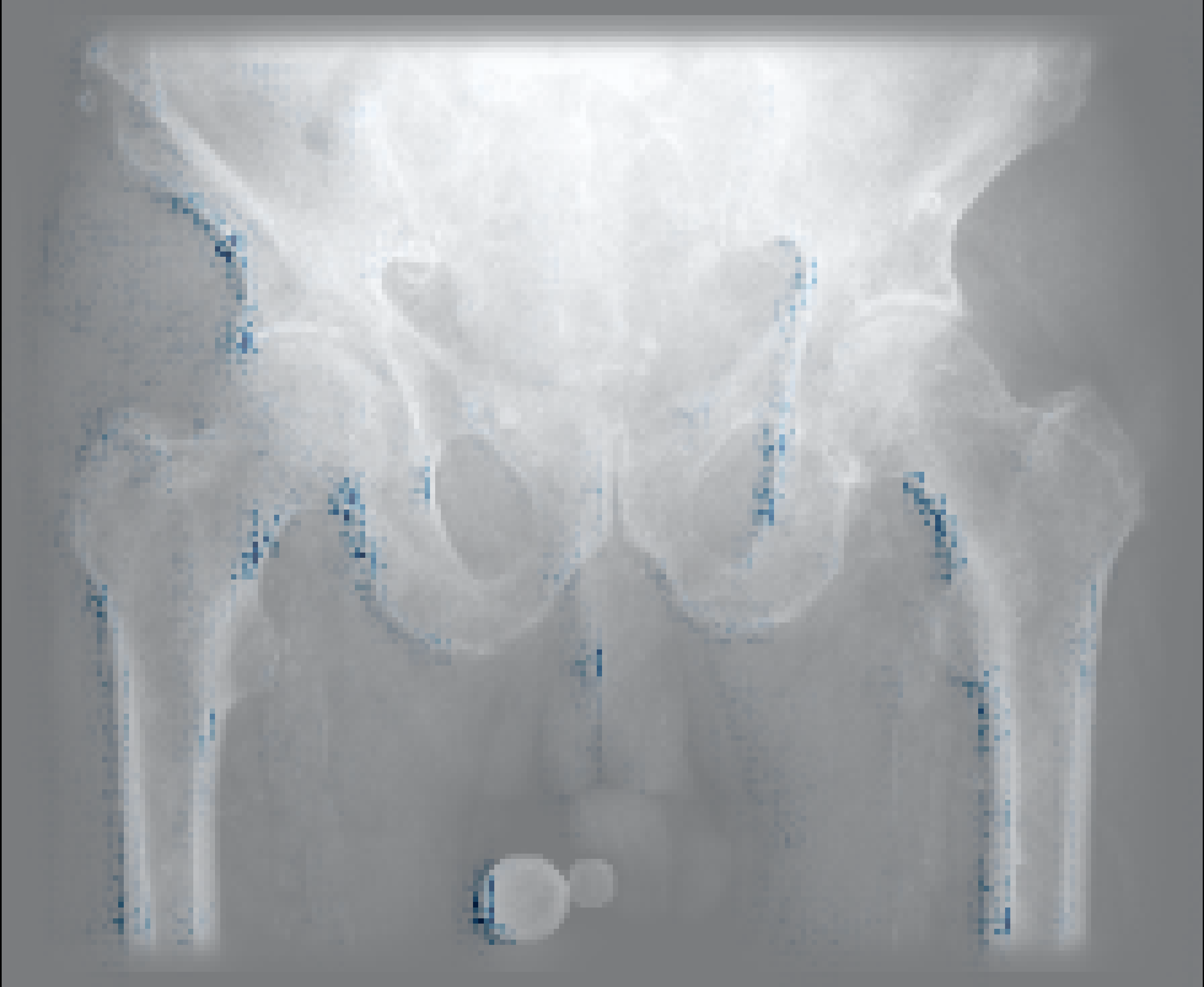}
        \caption{Blended}
        \label{subfig:GradCam:Blended}
    \end{subfigure}
    \begin{subfigure}[b]{\curfigwidth\textwidth}
        \includegraphics[width=\textwidth]{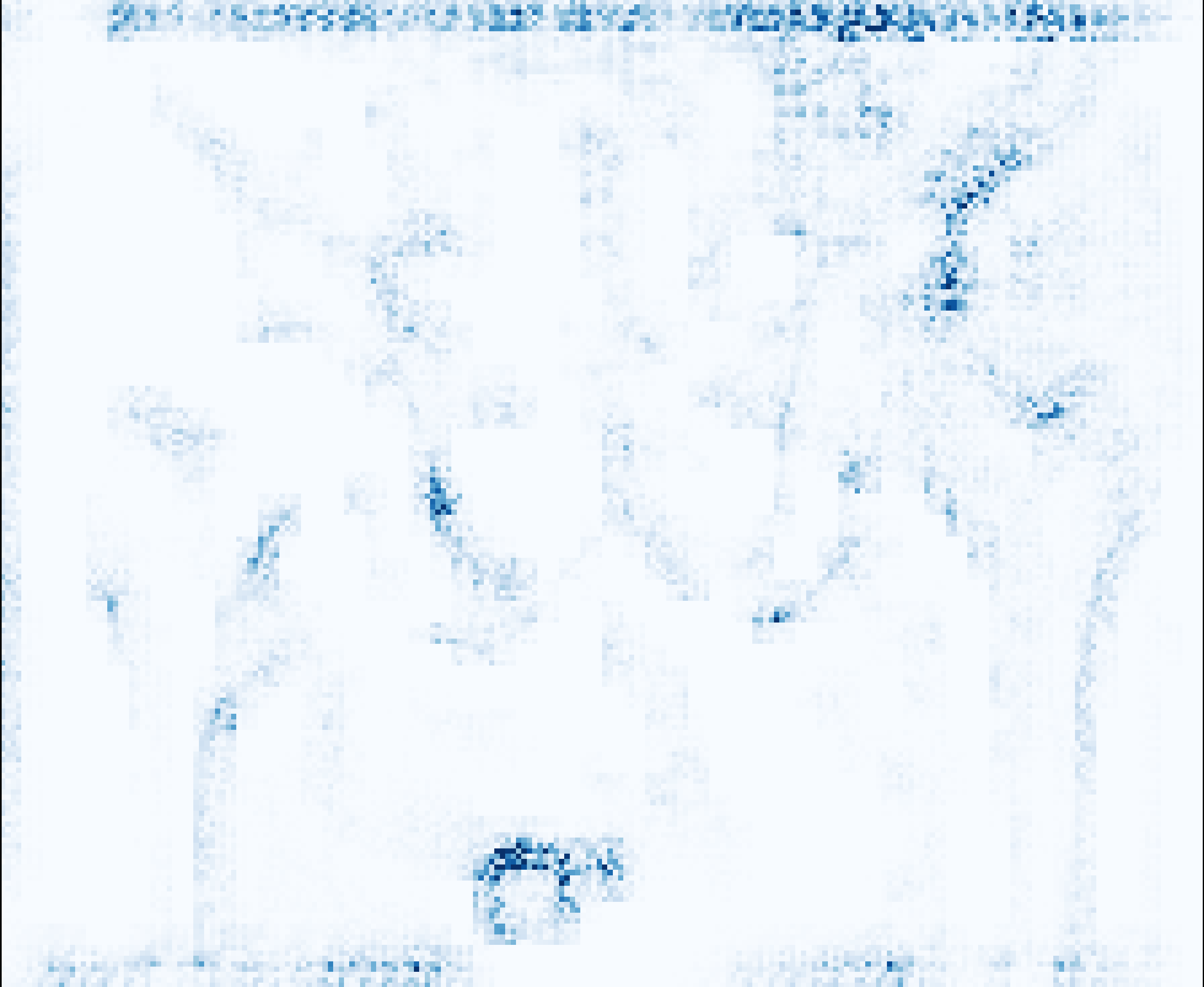}
        \includegraphics[width=\textwidth]{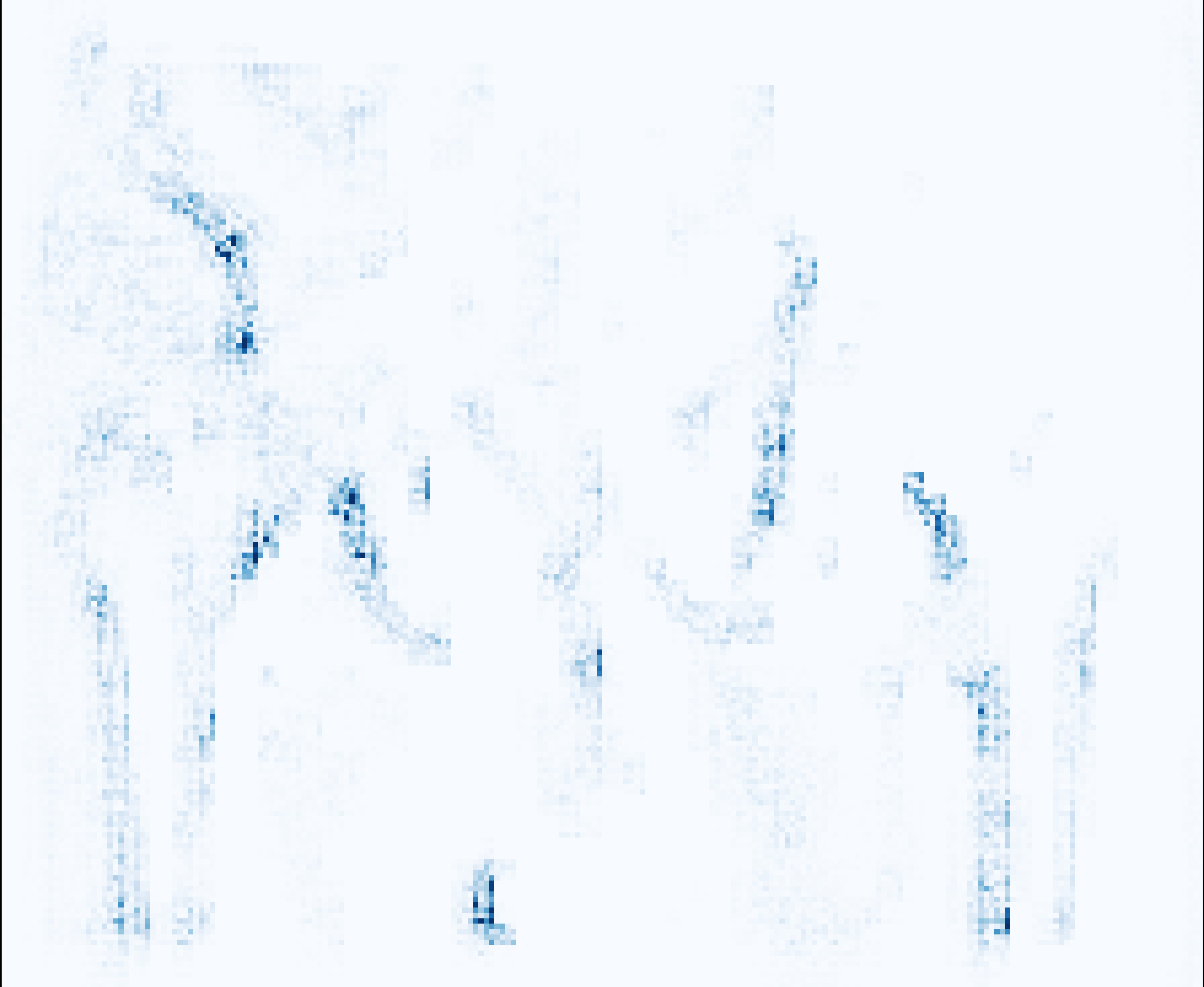}
        \caption{Heatmap}
        \label{subfig:GradCam:Heatmap}
    \end{subfigure}
    \caption{Guided GradCam \cite{Selvaraju2016GradCAMVE,Kokhlikyan2020CaptumAU} visualization of our SimCLR model trained without (top) or with (bottom) our custom data cleaning and augmentations enabled. Note the much stronger focus of the bottom model on medically relevant image regions, rather than the border and gauge.}
    \label{fig:GradCam}
\end{figure}

\subsection{Self-supervision excels in a low data regime}\label{subsec:Self-Supervision excels in a low data regime}
In \cref{fig:LowData}, we plot the effectiveness when training the final linear layer using our pretrained SimCLR/BYOL backbones with varying fractions of the training set -- SimCLR performs slightly better in very low data scenarios, yielding 92.2\% accuracy with labels using just 1\% of our training data (i.e. 310 images total across all 14 classes). A comparative baseline approach where we train a randomly initialized ResNet18 fully supervised with cross entropy on these subsets exhibits much steeper drops in performance, achieving just 57.1\% accuracy with 1\% of the labels.

\subsection{Cleaning+augmentation increase attention to anatomical areas}\label{subsec:Custom Augmentations and Plausibility}
To judge the impact of our custom data cleaning and gauge augmentation procedures described in \cref{subsec:Augmentation and Cleaning}, we utilize Guided GradCam \cite{Selvaraju2016GradCAMVE,Kokhlikyan2020CaptumAU} to perform a qualitative comparison on which parts of the image contribute more to the final prediction by following the gradient -- see \cref{fig:GradCam}. In terms of performance, there is very little overall impact on the overall accuracy ($\leq \pm 0.6\%$), but both for SimCLR and BYOL, the pelvis/hip class performance, for which we designed the gauge augmentation, increases by 2.3\% and 1.4\%, respectively.

\begin{table}[tb]
\caption{Additional evaluation results -- from top to bottom: Our ensemble approach on the original test set, the SimCLR model when manually filtering (F) and correcting images which were incorrectly labelled in our PACS  (see \cref{subsec:Data Quality Assessment}), and the performance of the ensemble on the same filtered and corrected set of labels. We observe increases of over $+3\%$ each for SimCLR in accuracy for the spine (s) classes after label corrections compared to \cref{tab:maintable}.}
\label{tab:secondarytable}
\begin{tabular}{@{}lrrrrrrrr@{}}
\toprule
\multicolumn{1}{l}{\textbf{Approach}} &
  \textbf{ALL} &
  ~\begin{tabular}[c]{@{}r@{}}clavicle\\      shoulder\end{tabular} &
  ~~~\begin{tabular}[c]{@{}r@{}}skull\\      rib\end{tabular} &
  ~~~\begin{tabular}[c]{@{}r@{}}elbow\\      knee\end{tabular} &
  ~~~\begin{tabular}[c]{@{}r@{}}wrist\\      hand\end{tabular} &
  ~~~\begin{tabular}[c]{@{}r@{}}foot\\      ankle\end{tabular} &
  ~\begin{tabular}[c]{@{}r@{}}pelvis/hip\\      cervical s\end{tabular} &
  ~\begin{tabular}[c]{@{}r@{}}thoracic s\\      lumbar s\end{tabular} \\ \midrule
\textbf{Ensemble} &
  97.7\% &
  \begin{tabular}[c]{@{}r@{}}98.6\%\\      98.7\%\end{tabular} &
  \begin{tabular}[c]{@{}r@{}}97.9\%\\      98.6\%\end{tabular} &
  \begin{tabular}[c]{@{}r@{}}97.7\%\\      99.1\%\end{tabular} &
  \begin{tabular}[c]{@{}r@{}}98.0\%\\      98.9\%\end{tabular} &
  \begin{tabular}[c]{@{}r@{}}98.7\%\\      98.7\%\end{tabular} &
  \begin{tabular}[c]{@{}r@{}}98.1\%\\      96.0\%\end{tabular} &
  \begin{tabular}[c]{@{}r@{}}92.3\%\\      95.6\%\end{tabular} \\ \midrule
\textbf{F SimCLR} &
  98.0\% &
  \begin{tabular}[c]{@{}r@{}}99.4\%\\      99.0\%\end{tabular} &
  \begin{tabular}[c]{@{}r@{}}100.0\%\\      98.2\%\end{tabular} &
  \begin{tabular}[c]{@{}r@{}}96.9\%\\      97.8\%\end{tabular} &
  \begin{tabular}[c]{@{}r@{}}97.6\%\\      99.3\%\end{tabular} &
  \begin{tabular}[c]{@{}r@{}}98.5\%\\      98.9\%\end{tabular} &
  \begin{tabular}[c]{@{}r@{}}95.5\%\\      98.5\%\end{tabular} &
  \begin{tabular}[c]{@{}r@{}}94.4\%\\      97.3\%\end{tabular} \\ \cmidrule(l){3-9} 
\textbf{F Ensemble} &
  98.8\% &
  \begin{tabular}[c]{@{}r@{}}99.8\%\\      99.6\%\end{tabular} &
  \begin{tabular}[c]{@{}r@{}}100.0\%\\      99.5\%\end{tabular} &
  \begin{tabular}[c]{@{}r@{}}98.2\%\\      99.3\%\end{tabular} &
  \begin{tabular}[c]{@{}r@{}}98.6\%\\      99.6\%\end{tabular} &
  \begin{tabular}[c]{@{}r@{}}99.8\%\\      99.4\%\end{tabular} &
  \begin{tabular}[c]{@{}r@{}}96.4\%\\      98.3\%\end{tabular} &
  \begin{tabular}[c]{@{}r@{}}95.4\%\\      98.1\%\end{tabular} \\ \bottomrule
\end{tabular}
\end{table}

\begin{figure}[tb]
    \centering
    \begin{subfigure}[b]{0.49\textwidth}
        {\includegraphics[width=\textwidth]{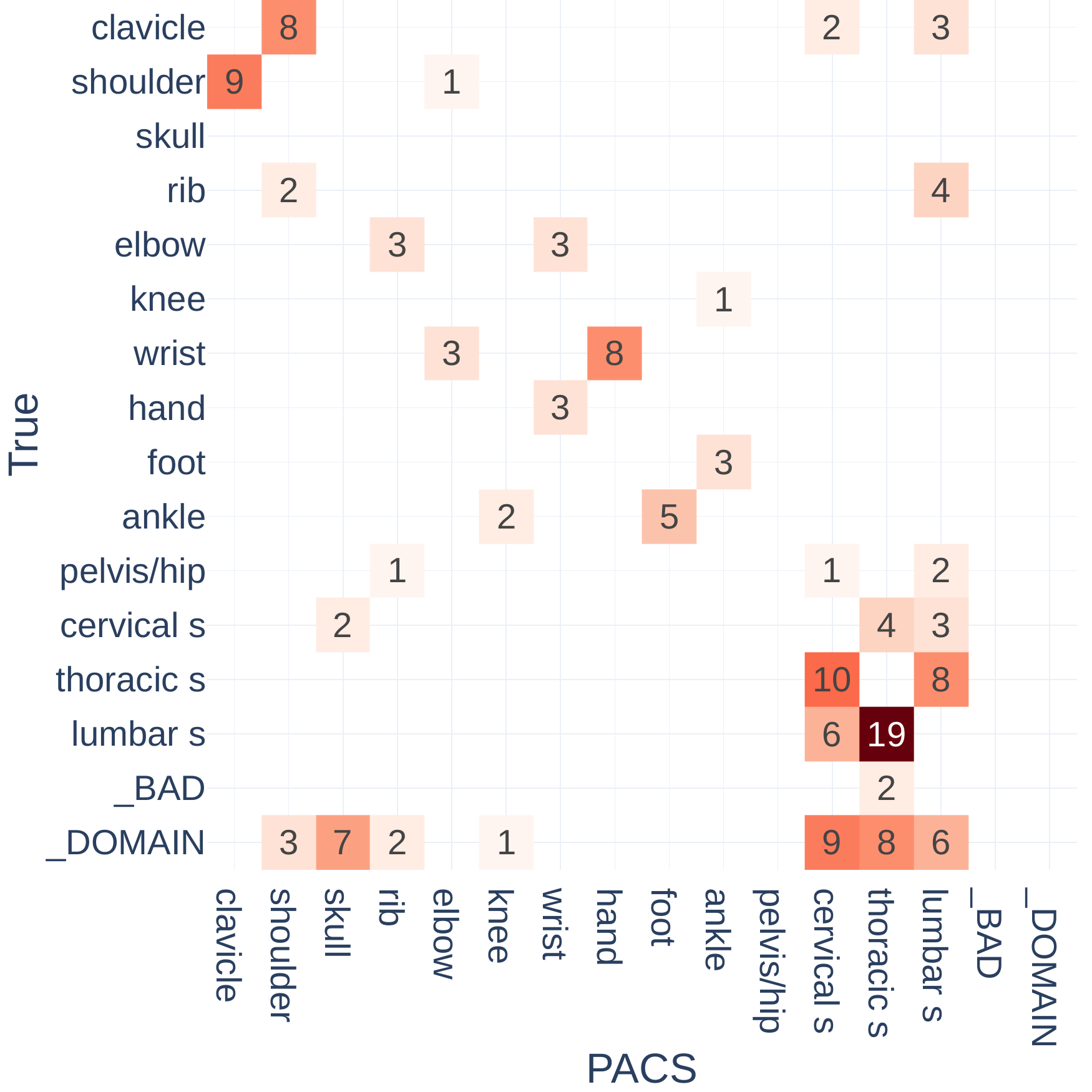}}
        \captionsetup{width=.95\linewidth}
        \caption{Non-diagonal elements of the confusion matrix (CM) between the old PACS labels and our new ones: note the discrepancies in incorrect spine label assignments, as well as clavicle vs. shoulder and hand vs. wrist. Also, two images are discarded due to bad quality and a further 36 due to being out-of-domain.\\~}
        \label{subfig:QualityAssessment:PACS}
    \end{subfigure}
    \begin{subfigure}[b]{0.49\textwidth}
        {\includegraphics[width=\textwidth]{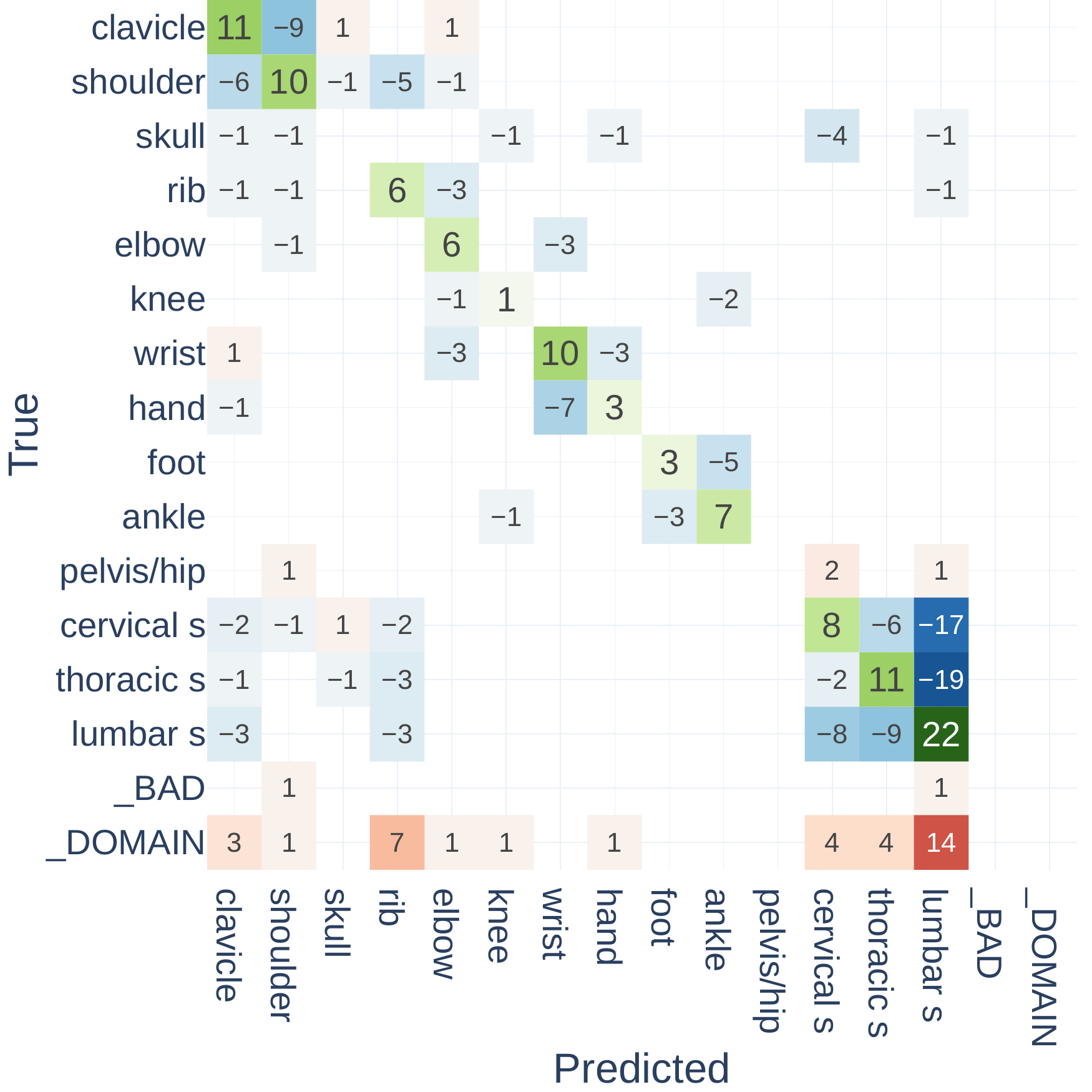}}
        \captionsetup{width=.95\linewidth}
        \caption{Difference (delta) $CM_{new} - CM_{old}$ for our SimCLR model: the green elements on the diagonal are additional correct predictions (total 98), blue (negative value) entries are no longer present ``incorrect'' predictions, and red (positive value) represent new incorrect predictions. Note that almost all new incorrect predictions are out-of-domain.
        }
        \label{subfig:QualityAssessment:Main}
    \end{subfigure}
    \caption{Confusion matrices (CM) describing the results of our quality assessment on SimCLR's ``incorrect'' predictions (see \cref{subsec:Data Quality Assessment}). In both cases, the rows refer to our new expert radiologist's labels.}
    \label{fig:QualityAssessment}
\end{figure}

\subsection{Ensembles further increase performance}\label{subsec:Ensemble}
In order to further boost the performance of our approach, we build an ensemble from our 3 primary architectures in \cref{tab:maintable} by adding their outputs after applying softmax to the logits, and then selecting the class with the highest score -- this yields an overall accuracy of 97.7\% (\cref{tab:secondarytable}).

\subsection{Detecting and fixing noisy PACS labels}\label{subsec:Data Quality Assessment}
Since our ``ground-truth'' labels were directly scraped from PACS, they are not perfectly reliable. To analyze their quality, we tasked an expert radiologist with re-evaluating all test failure cases of the SimCLR model (n=328), almost half of which turned out to be either incorrectly labeled (n=116), out-of-domain (i.e. not in the 14 classes) (n=36), or of unusable quality (n=2) (\cref{fig:QualityAssessment}). A total of 98 of the 116 incorrectly labeled in-domain images are in fact correctly predicted by our SimCLR model, increasing its theoretical accuracy (when excluding out-of-domain and unusable images) to 98.0\%, and the ensemble's to 98.8\% (\cref{tab:secondarytable}).

Due to the very high quality of our predictions, we employ them to detect incorrect anatomical region candidates in PACS metadata, which can subsequently be checked or preemptively skipped for future research endeavors, improving the real-life clinical dataset consistency. Since our architecture only fits a single fully connected layer using the noisy PACS labels, it is feasible to apply the inconsistency detection to images which were part of the training/validation set in the future.

In our observations, especially the anatomical region annotations on the spine and shoulder vs. clavicle are prone to inconsistencies -- see both \cref{fig:QualityAssessment} and our T-SNE \cite{Maaten2008VisualizingDU} plot in the supplementary material; this can be explained by the overlap of several anatomical regions within a small examination area.

\section{Conclusion}
In this paper, we introduced our -- ready for real-world application -- self-super\discretionary{-}{}{}vised architecture for radiograph anatomical region classification. Due to its blend of flexibility w.r.t.~label availability ($\approx 300$ labeled instances) and high accuracy ($96.6\% - 98.8\%$ accuracy), it can be utilized \textsc{a)} to increase the quality of existing metadata collections, \textsc{b)} improve downstream application reliability and research dataset sizes, as well as \textsc{c)} generate PACS metadata in low-resource scenarios from scratch. The training is further complemented by our custom augmentation techniques. 

Demonstrating our method's effectiveness in finding inconsistencies despite learning from the same noisy ground-truth source, an expert radiologist successfully identified over 150 incorrect PACS labels, after analyzing a subset of 328 candidates generated by our method. This research aims at sparking interest in applying such methods to real-world clinical scenarios.

\clearpage{}

\bibliographystyle{splncs04}
\bibliography{paper}

\newpage
\title{\normalsize Self-Supervised Radiograph Anatomical Region Classification -- \\How Clean Is Your Real-World Data?\\(Supplementary Material)}
\titlerunning{Self-Supervised Radiograph Anatomical Region Classification}
\author{}

\authorrunning{S. Langer et al.}

\institute{}
\maketitle              

\begin{figure}[h!]
    \centering
    \begin{subfigure}[b]{0.24\textwidth}
        \includegraphics[width=\textwidth]{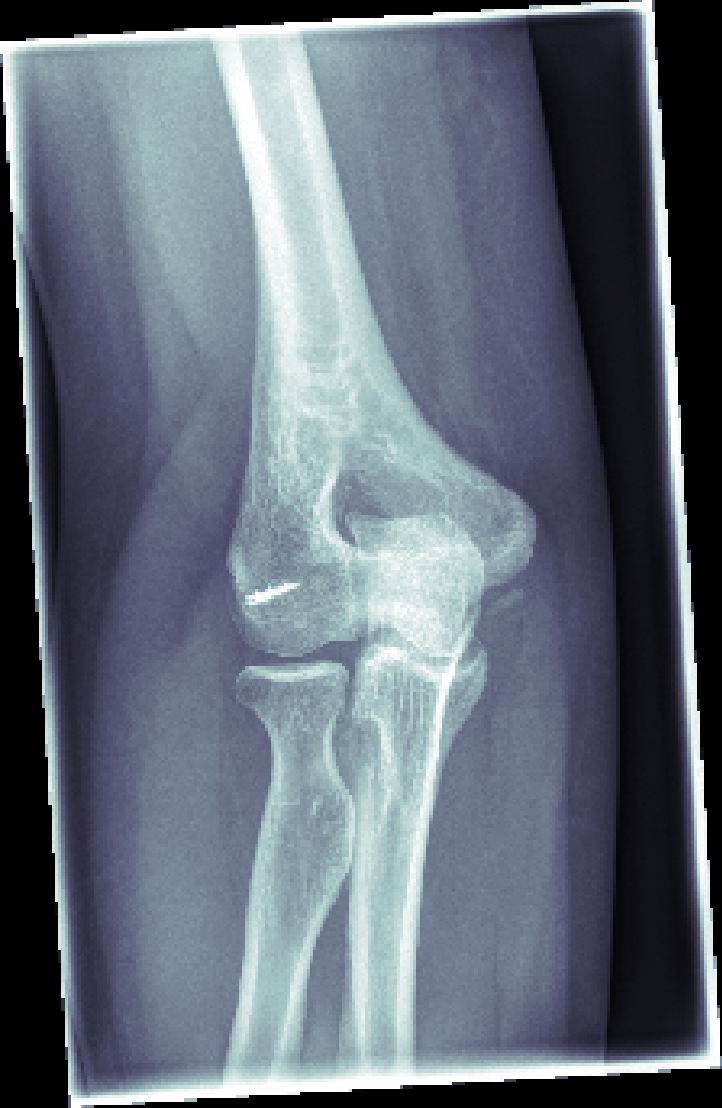}
        \caption{Raw Image}
        \label{subfig:Augmentation:Raw Image}
    \end{subfigure}
    \begin{subfigure}[b]{0.24\textwidth}
        \includegraphics[width=\textwidth]{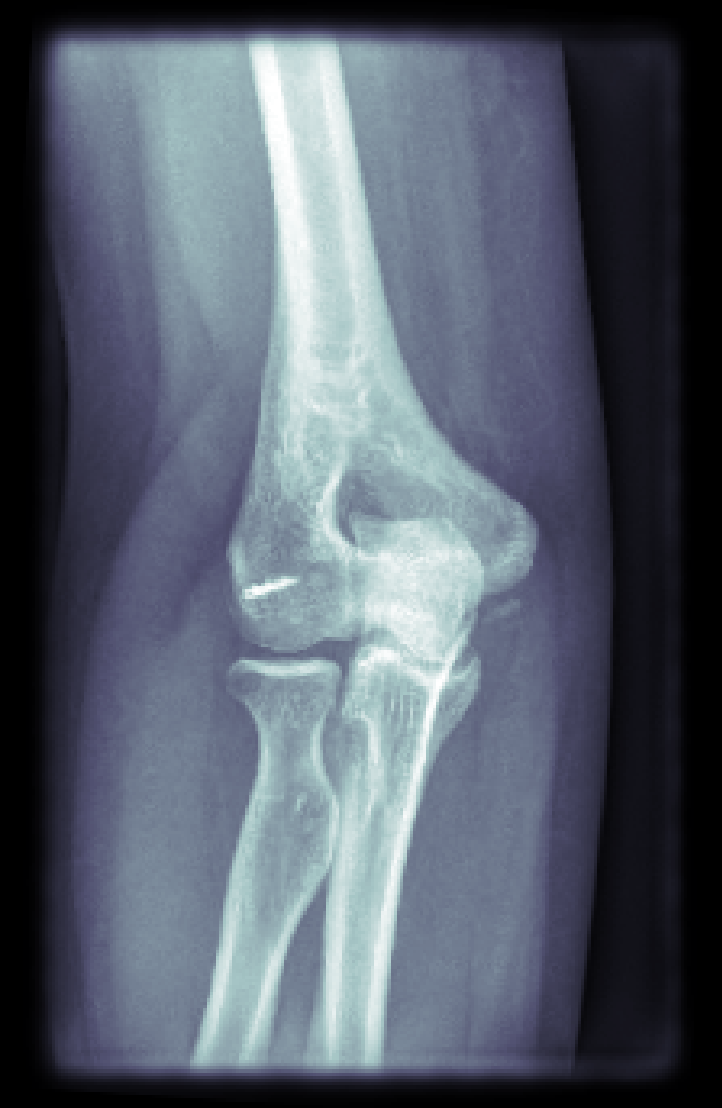}
        \caption{Preprocessed}
        \label{subfig:Augmentation:Preprocessed}
    \end{subfigure}
    \begin{subfigure}[b]{0.24\textwidth}
        \includegraphics[width=\textwidth]{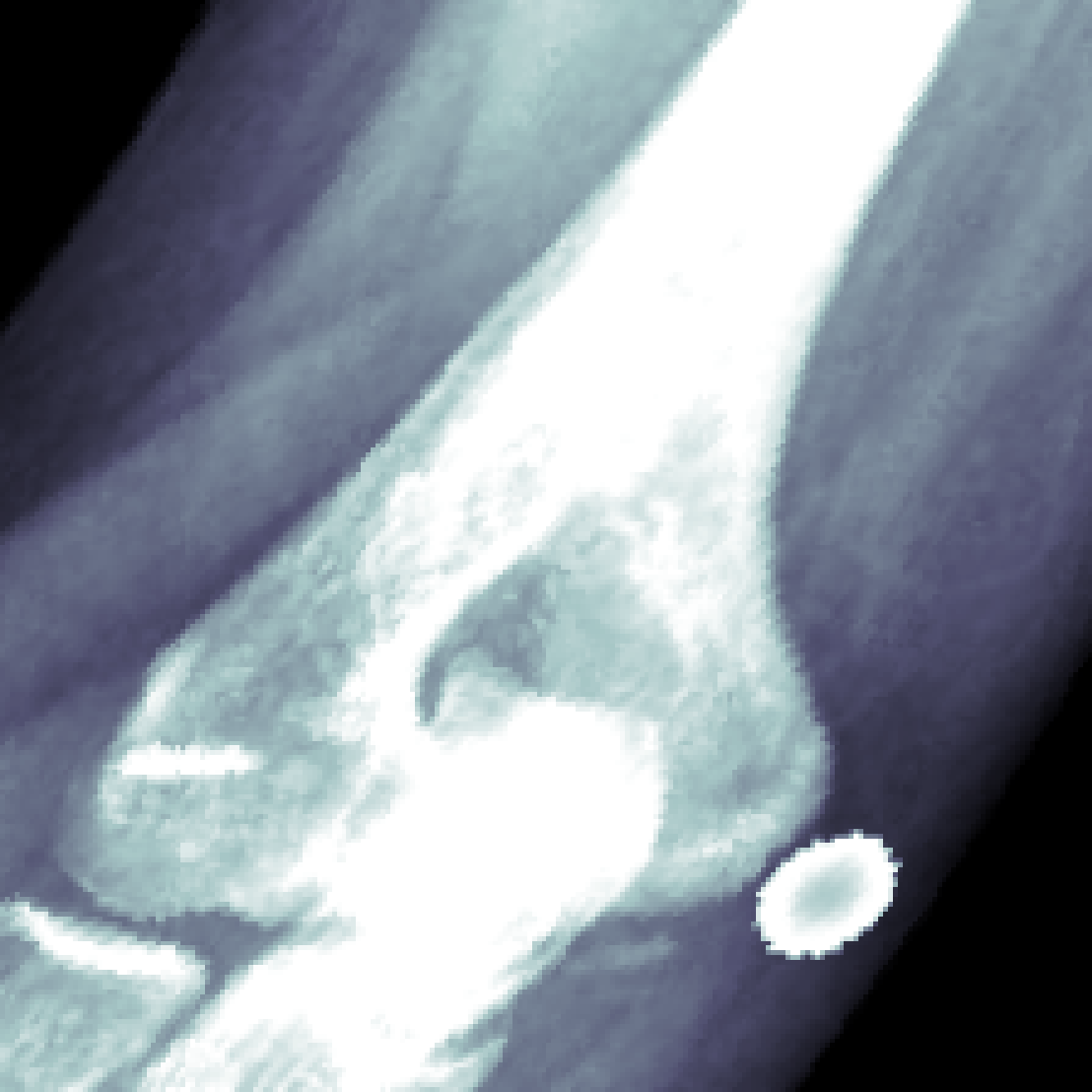}
        \captionsetup{justification=centering}
        \caption{Augmented \\for Pre-training}
        \label{subfig:Augmentation:Pretraining}
    \end{subfigure}
    \begin{subfigure}[b]{0.24\textwidth}
        \includegraphics[width=\textwidth]{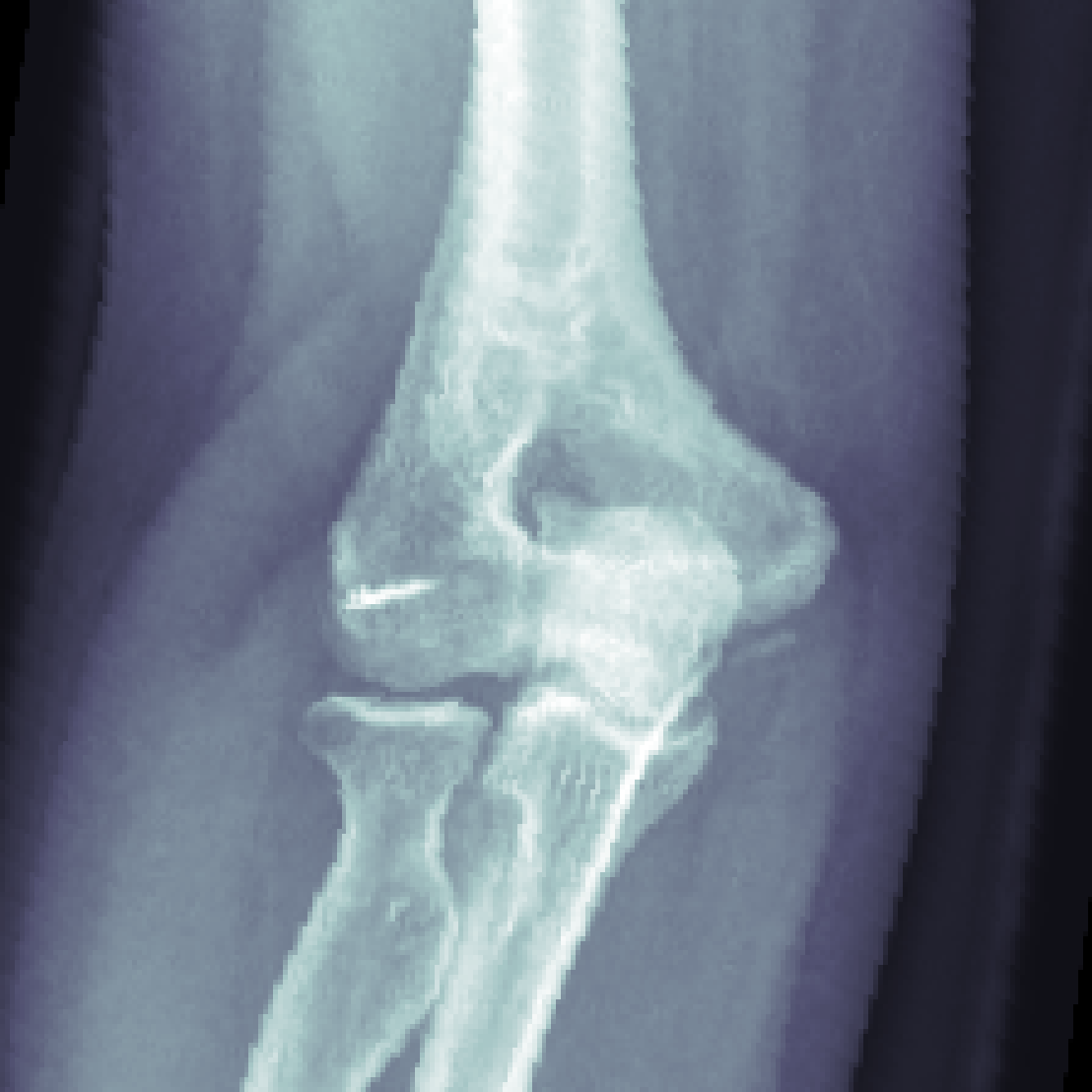}
        \captionsetup{justification=centering}
        \caption{Augmented \\for Training}
        \label{subfig:Augmentation:Finetuning}
    \end{subfigure}
    \caption{Evolution of an image -- the strengths of color jitter, random affine and random resized crop augmentations are noticeably higher during the self-supervised pretraining in order to ensure the extraction of relevant features.}
    \label{fig:Augmentation}
\end{figure}

\begin{figure}[h!]
    \centering
    \includegraphics[width=\textwidth]{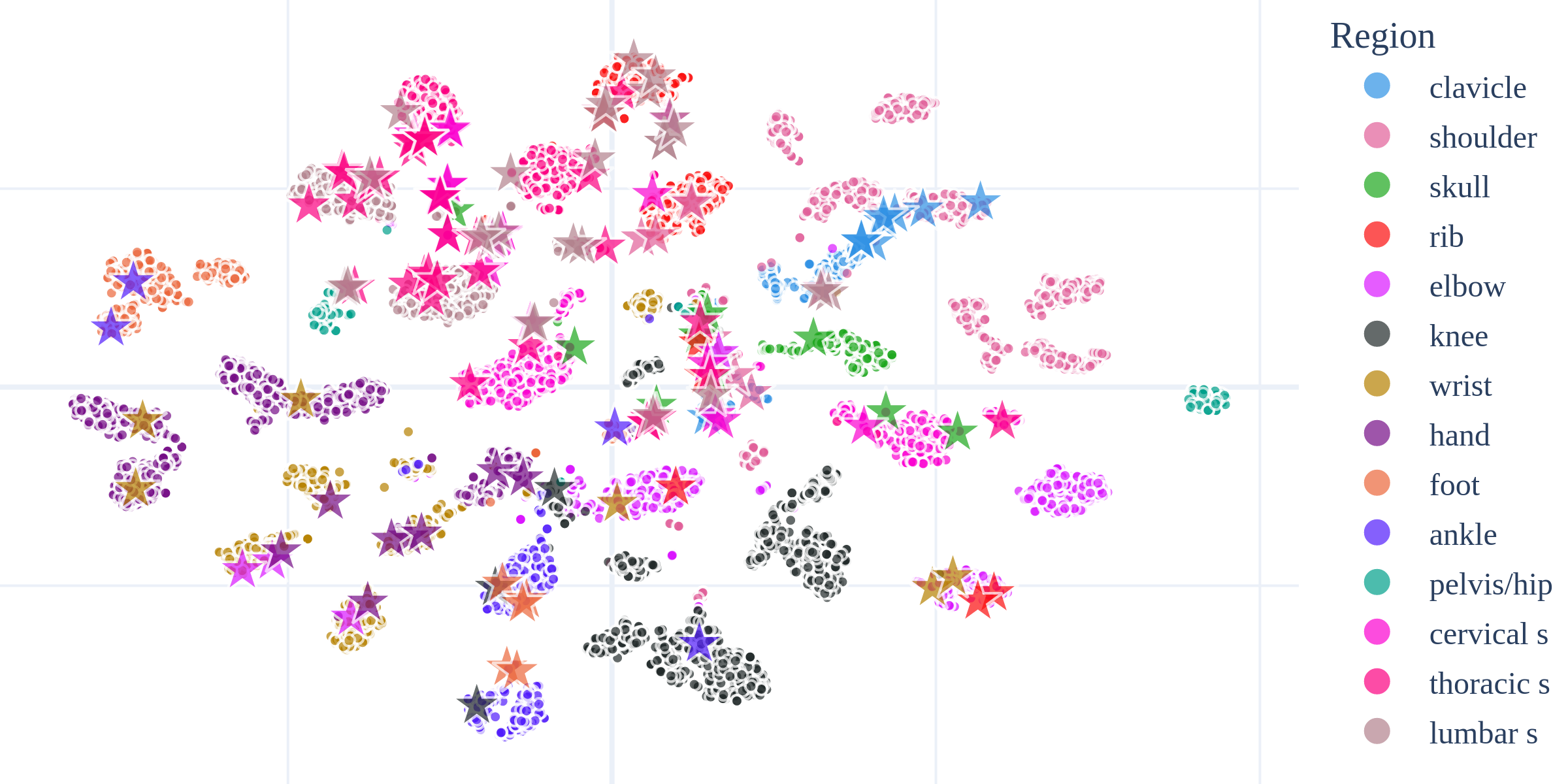}
    \caption{T-SNE visualization of the test images' features, generated by our SimCLR pre-training -- PACS labels later identified as incorrect have a star marker.}
    \label{fig:TSNE}
\end{figure}

\begin{table}[htb]
\caption{Overview of our augmentation hyperparameters.}
\label{tab:hyperparams}
\begin{tabularx}{\textwidth}{XXX}
\toprule
\textbf{Hyperparameter} & \textbf{Pre-training}  & \textbf{Training} \\ \midrule
Gauge Occurrences       & {$\frac{1}{3}$ probability for 0, 1, 2} & {$\frac{1}{3}$ probability for 0,1,2} \\
Gauge Scale             & $[0.8, 1.2]$ & $[0.8, 1.2]$          \\
Gauge Opacity           & $[0.75, 1]$ & $[0.75, 1]$           \\
Color Jitter Brightness & $[0.5, 1.5]$        & $[0.9, 1.1]$      \\
Color Jitter Contrast   & $[0.25, 1.75]$      & $[0.8, 1.2]$      \\
Affine Rotation         & $[-15, 15]$         & $[-5, 5]$         \\
Affine Translation      & $[-0.1, 0.1]$       & $[-0.02, 0.02]$   \\
Affine Scale            & $[0.8, 1.5]$        & $[0.95, 1.1]$     \\
Affine Shear            & $[-30, 30]$         & $[-10, 10]$       \\
Random Resize Scale     & $[0.08, 1]$         & $[0.95, 1.1]$     \\
Random Resize Ratio     & $[\frac{3}{4}, \frac{4}{3}]$        & $[0.9, 1.1]$      \\ \bottomrule
\end{tabularx}
\end{table}

\begin{table}[htb]
\caption{Number of radiographs per anatomical region and dataset split.}
\label{tab:datasplit}
\begin{tabularx}{\textwidth}{rXXX}
\toprule
\textbf{Anatomical Region} & \hfill \textbf{training} & \hfill \textbf{validation} & \hfill \textbf{test} \\ \midrule
\textbf{clavicle}          & \hfill 1456              & \hfill 369                 & \hfill 499           \\
\textbf{shoulder}          & \hfill 3953              & \hfill 1023                & \hfill 1164          \\
\textbf{skull}             & \hfill 728               & \hfill 178                 & \hfill 280           \\
\textbf{rib}               & \hfill 1743              & \hfill 430                 & \hfill 552           \\
\textbf{elbow}             & \hfill 2293              & \hfill 521                 & \hfill 809           \\
\textbf{knee}              & \hfill 3616              & \hfill 946                 & \hfill 1165          \\
\textbf{wrist}             & \hfill 2938              & \hfill 701                 & \hfill 830           \\
\textbf{hand}              & \hfill 3030              & \hfill 765                 & \hfill 1002          \\
\textbf{foot}              & \hfill 1405              & \hfill 329                 & \hfill 466           \\
\textbf{ankle}             & \hfill 1762              & \hfill 452                 & \hfill 540           \\
\textbf{pelvis/hip}        & \hfill 741               & \hfill 167                 & \hfill 216           \\
\textbf{cervical s}        & \hfill 2752              & \hfill 664                 & \hfill 885           \\
\textbf{thoracic s}        & \hfill 2022              & \hfill 508                 & \hfill 608           \\
\textbf{lumbar s}          & \hfill 2512              & \hfill 684                 & \hfill 730           \\ \bottomrule
\end{tabularx}
\end{table}

\end{document}